\documentclass[fleqn,10pt]{wlscirep}
\usepackage[utf8]{inputenc}
\usepackage[T1]{fontenc}
\usepackage{hyperref}
\usepackage{graphics}
\usepackage{epsfig}
\usepackage{xspace}
\usepackage{wrapfig,enumitem}
\usepackage{multirow}
\usepackage{subcaption}
\usepackage{svg}

\title{A comprehensive multimodal dataset and benchmark for ulcerative colitis scoring in endoscopy}
\author[1]{Noha Ghatwary}
\author[2]{Jiangbei Yue}
\author[3]{Ahmed Elgendy}
\author[4]{Hanna Nagdy}
\author[5]{Ahmed Galal}
\author[6]{Hayam Fathy}
\author[6]{Hussein El-Amin}
\author[7]{Venkataraman Subramanian}
\author[7]{Noor Mohammed}
\author[8]{Gilberto Ochoa-Ruiz}
\author[2,*]{Sharib Ali}

\affil[1]{Computer Engineering Department, Arab Academy for Science and Technology,Smart Village, Giza, Egypt}
\affil[2]{AI in Medicine and Surgery group, School of Computer Science, University of Leeds, LS2 9JT, Leeds, United Kingdom}
\affil[3]{Department of Electrical and Computer Engineering, Queen's University, Canada}
\affil[4]{Internal Medicine Department, College of Medicine, Arab Academy for Science and Technology, Egypt}
\affil[5]{Internal medicine, Alexandria University, Egypt}
\affil[6]{Internal medicine, Division Hepatogastroenterology, Assiut university, Egypt}
\affil[7]{Leeds Teaching Hospital NHS Trust, Leeds, United Kingdom}
\affil[8]{School of Engineering and Sciences, Tecnologico de Monterrey, Monterrey, Mexico}
\affil[*]{corresponding author: Sharib Ali (s.s.ali@leeds.ac.uk)}
 
\begin{abstract}
Ulcerative colitis (UC) is a chronic mucosal inflammatory condition that places patients at increased risk of colorectal cancer. Colonoscopic surveillance remains the gold standard for assessing disease activity, and reporting typically relies on standardised endoscopic scoring metrics. The most widely used is the Mayo Endoscopic Score (MES), with some centres also adopting the Ulcerative Colitis Endoscopic Index of Severity (UCEIS). Both are descriptive assessments of mucosal inflammation (MES: 0–3; UCEIS: 0–8), where higher values indicate more severe disease.
However, computational methods for automatically predicting these scores remain limited, largely due to the lack of publicly available expert‑annotated datasets and the absence of robust benchmarking. There is also a significant research gap in generating clinically meaningful descriptions of UC images, despite image captioning being a well‑established computer vision task. Variability in endoscopic systems and procedural workflows across centres further highlights the need for multi‑centre datasets to ensure algorithmic robustness and generalisability.
In this work, we introduce a curated multi‑centre, multi‑resolution dataset that includes expert‑validated MES and UCEIS labels, alongside detailed clinical descriptions. To our knowledge, this is the first comprehensive dataset that combines dual scoring metrics for classification tasks with expert‑generated captions describing mucosal appearance and clinically accepted reasoning for image captioning. This resource opens new opportunities for developing clinically meaningful multimodal algorithms. In addition to the dataset, we also provide benchmarking using convolutional neural networks, vision transformers, hybrid models, and widely used multimodal vision‑language captioning algorithms.

\end{abstract}
\begin{document}

\flushbottom
\maketitle
\thispagestyle{empty}
\section*{Background}
Ulcerative colitis (UC) is a chronic intestinal inflammatory disease that primarily affects the colon and rectum. Patients with long-standing UC have an elevated risk of developing colorectal cancer (CRC). Globally, CRC remains a major health burden, causing approximately 930,000 deaths~\cite{Rabeneck2020} per year and ranking as the second leading cause of cancer-related mortality worldwide. UC significantly increases the long‑term risk of developing CRC (2–4‑fold higher risk) due to chronic, persistent inflammation of the colonic mucosa. 

Early identification and monitoring of ulcerative colitis (UC) are essential to prevent disease progression, complications, and long‑term colorectal cancer risk. Colonoscopy remains the gold‑standard modality for diagnosing UC, assessing the disease extent, and conducting ongoing endoscopic surveillance, given its ability to directly visualize mucosal inflammation and obtain tissue biopsies.  Multiple endoscopic indices have been developed to standardize the evaluation of UC severity; among these, the Mayo Endoscopic Score (MES) is consistently recognized as one of the most widely applied and guideline‑recommended scoring systems in both clinical practice and clinical research~\cite{Kochhar2025}.  Its widespread adoption reflects its practicality, simplicity, and strong familiarity among gastroenterologists, who routinely use MES to quantify inflammatory activity and guide therapeutic decision‑making.
Comparative evaluations demonstrate that MES is more widely recognized and utilized than other indices such as the Ulcerative Colitis Endoscopic Index of Severity (UCEIS), reflecting its practical simplicity and broad adoption in gastroenterology practice. However, in a study comparing the MES and UCEIS scoring~\cite{Travis2012-hl}, the authors concluded that the UCEIS accurately reflects clinical outcomes and predicted the medium- to long-term prognosis in UC patients undergoing induction therapy~\cite{Ikeya2016-fz}.   

The MES score is traditionally characterized and graded according to morphological information: i.e., vascular patterns, the mucosal topology, structural rectal bleeding, among other factos. However, the grading and interpretation of UC in colonoscopy highly depends on the clinician's experience level. Although experts can assign follow-up scores, differences in colonoscopy assessments can still affect patient diagnosis. For these reasons, the development of computer-aided diagnosis tools based on computer vision is of paramount importance to assist the work of the gastroenterologists during the UC examinations and to improve the patients outcomes.

During the past decade, advances in computer vision and deep learning have accelerated the automation of endoscopic analysis, showing substantial progress in lesion detection, segmentation, and disease classification \cite{ref_cv_dl_endoscopy}.  Convolutional neural networks (CNN) have shown reliable performance in identifying ulcers and vascular changes, while more recently transformer-based architectures have brought greater representation power and contextual awareness \cite{ref_vlm_medical}. However, these CADx methods largely focus on classification and detection tasks, providing limited insight into the semantic or relational context of what is being observed. Their outputs are numeric or categorical, offering little transparency into how or why a specific decision was made. As a result, their clinical adoption remains constrained by a persistent gap between model accuracy and model interpretability.

Recently, the emergence of vision language models (VLMs) has opened a promising new avenue for understanding medical images \cite{ref_vlm_medical}. By coupling visual encoders with language decoders, these architectures can generate descriptive sentences, "captions", that articulate image content in natural language. In the medical domain, this capacity holds particular promise: a well-constructed caption can simultaneously communicate the visual features of an image and the clinical reasoning behind a diagnosis. However, despite this potential, the application of captioning in medical imaging remains in its infancy. 

Parallel to these developments, explainable artificial intelligence (XAI) has become a central theme in biomedical computing \cite{ref_xai_biomedical}. Methods such as Grad-CAM and attention visualization have provided post hoc interpretations of model behavior, highlighting regions that contribute most to a prediction. While valuable, such explanations are inherently reactive---they attempt to rationalize decisions after they occur rather than embedding interpretability directly within the reasoning process. For medical imaging, this distinction is critical: clinicians require systems that do not merely ``show where the model looked,'' but rather ``understand what the model saw'' in relation to anatomical or pathological structures. Bridging this cognitive gap demands architectures that unify visual understanding, relational reasoning, and language generation within a transparent computational framework. 
This limitation becomes particularly evident in clinical domains that require explicit evidence-based interpretation, such as ulcerative colitis endoscopy.

From a clinical research perspective, the need for interpretable and standardized assessment in ulcerative colitis has motivated a growing body of work on computer-aided analysis of endoscopic imagery. Early approaches focused on image-level severity classification and lesion detection, demonstrating that deep learning models can capture discriminative visual patterns associated with mucosal inflammation. More recent efforts have explored attention mechanisms and weak localization to improve transparency, highlighting regions that contribute most to disease grading decisions. However, despite these advances, most existing systems remain centered on prediction accuracy and provide limited support for explicit clinical reasoning or narrative interpretation.

 Deep learning (DL) based automated UC grading methods could help reduce the operator subjectivity observed in complex UC scoring and improve diagnostic quality. Even though some DL methods have been devised, these methods 1) have below 80\% classification accuracy,  2) must be rigorously assessed on multi-centre data 3) must include language descriptions to improve the model interpretability. 

 Despite the rapid progress in computer vision and deep learning for endoscopic analysis, the translation of these methods into reliable clinical tools remains limited. A key obstacle is the scarcity of large, diverse, and well-annotated multi-center datasets that capture the variability inherent in real-world clinical practice, including differences in endoscopic equipment, imaging protocols, patient populations, and disease presentation. Models trained on limited or homogeneous datasets often fail to generalize across institutions, which restricts their clinical applicability. Moreover, the lack of standardized benchmarks makes it difficult to objectively compare competing methods and assess their robustness, interpretability, and clinical relevance. Addressing these challenges requires collaborative efforts to curate diverse datasets and establish open evaluation frameworks. In this context, community-driven challenges play a critical role by providing shared datasets, standardized evaluation protocols, and transparent comparisons of algorithms. Such initiatives can accelerate methodological innovation, foster reproducibility, and ultimately support the development of computer-aided diagnostic systems that are robust, interpretable, and clinically deployable for ulcerative colitis assessment.

\begin{table*}[t]
\centering
\caption{Distribution of the MES and UCEIS scores for the full dataset comprised of 2945 images total from the three Subsets. Each value indicate the number of images in each score combination.}
\label{tab:mes_uceis_totals}

\resizebox{\textwidth}{!}{
\begin{tabular}{c|ccccccccc|c}
\hline
\textbf{} 
& \textbf{UCEIS-0} & \textbf{UCEIS-1} & \textbf{UCEIS-2} & \textbf{UCEIS-3} 
& \textbf{UCEIS-4} & \textbf{UCEIS-5} & \textbf{UCEIS-6} 
& \textbf{UCEIS-7} & \textbf{UCEIS-8} 
& \textbf{Total} \\
\hline
MES-0 & 285 & 6 & 3 & 0 & 0 & 0 & 0 & 0 & 0 & 294 \\
MES-1 & 4 & 202 & 252 & 45 & 15 & 5 & 1 & 0 & 0 & 524 \\
MES-2 & 1 & 0 & 129 & 319 & 254 & 124 & 81 & 0 & 0 & 908 \\
MES-3 & 0 & 0 & 2 & 7 & 24 & 221 & 173 & 150 & 192 & 769 \\
\hline
\textbf{Total} 
& 290 & 208 & 386 & 371 & 293 & 350 & 255 & 150 & 192 & 2495 \\
\hline
\end{tabular}
}

\end{table*}

In this paper we present introduce a comprehensive multi-modal dataset and benchmarking for ulcerative colitis scoring in endoscopy. The results presented herein provide a common benchmark that will enable other researchers to evaluate and compare algorithms under the same conditions. The dataset and benchmark will promote methodological transparency, and help to accelerate innovation in clinically relevant tasks such as disease grading and captioning. To address this need, the benchmark was designed through a collaborative effort between clinicians and computer vision researchers. A curated dataset of colonoscopic images was collected and carefully annotated by expert gastroenterologists following established clinical scoring systems. The data were then divided into training, validation, and test sets to ensure unbiased evaluation. Clear evaluation protocols and metrics were defined to assess algorithm performance, enabling participants to develop and test AI models capable of assisting clinicians in the objective and reproducible assessment of ulcerative colitis severity.

\section*{Related work}
Several studies have proposed deep learning approaches for automated grading of ulcerative colitis (UC) based on colonoscopic images. 
Sutton et al.~\cite{sutton2022artificial} employed a DenseNet121 architecture and reported an accuracy of 87.5\% with an AUC of 0.90 on the HyperKvasir dataset. Similarly, Ozawa et al.~\cite{ozawa2019novel} developed a computer-aided diagnostic system based on GoogLeNet trained on 26,304 colonoscopy images collected from 841 patients, achieving an AUROC of 0.98 when distinguishing between Mayo scores 0–1 and 2–3. However, both approaches focused on binary classification tasks.

To address more detailed grading, Bhambhvani and Zamora \cite{bhambhvani2021deep} proposed a ResNeXt-101 model capable of discriminating individual Mayo Endoscopic Scores (MES 1, 2, and 3). Their method obtained AUC values of 0.96, 0.89, and 0.86 for MES 3, 2, and 1 respectively, with an overall accuracy of 77.2\%. In a similar line of work, Xu et al. \cite{xu2022additive} introduced a three-class MES classification model based on the EfficientNet-b5 architecture using the HyperKvasir dataset. Their approach incorporated ArcFace loss, an additive angular margin penalty applied to the softmax function, and achieved a top-1 accuracy of 75.06

Despite these advances, existing methods focus solely on classification and do not incorporate image captioning mechanisms to generate textual descriptions of endoscopic findings.

To date, relatively few studies have explored image captioning techniques in endoscopic diagnosis. Fonollà et al. \cite{fan2022novel} proposed a computer-aided diagnosis (CADx) system designed to automatically generate reports for colorectal polyps (CRP) according to the Blue Light Imaging Adenoma Serrated International Classification (BASIC), using four descriptive features. Their model employs an EfficientNetB4 architecture as an encoder and visual feature extractor, where the original classification layers are replaced with a global average pooling layer. For textual modeling, the authors incorporated a pre-trained BERT module to learn polyp-related semantic representations. Visual and textual features are subsequently fused and processed by an LSTM network to model temporal dependencies within the generated descriptions. The system was evaluated using standard $n$-gram-based captioning metrics, including BLEU, ROUGE, and METEOR. The model achieved a BLEU-1 score of 0.67, a ROUGE-L score of 0.83, and a METEOR score of 0.50. Despite these promising results, to the best of our knowledge, no prior work has addressed the automatic generation of clinical reports for ulcerative colitis using image captioning techniques.

In other areas of medical image analysis, captioning has been explored across several clinical domains where visual interpretation plays a decisive diagnostic role. In radiology, Jinh et al. \cite{jing2018} pioneered automated report generation using convolutional--recurrent networks, while Li et al.  \cite{li2020} compared hierarchical transformers to recurrent approaches for chest X-ray reporting. More recently, Xue et al. \cite{xue2022} advanced this line by integrating contextual attention to improve coherence, and Yuan et al. \cite{yuan2021automatic} demonstrated that transformer-based encoders could achieve human-like phrasing in chest X-ray descriptions. More recently, Liu et al. \cite{liu2023}  proposed a unified multi-modal framework that captures both image and structured metadata for clinical reporting. Beyond radiology, pathology has seen growing interest through hierarchical or fine-grained captioning approaches such as PathCap \cite{wang2021pathcap} , the attention-based HistoCap \cite{mitra2023histocap} , and the histological captioning model of He et al. \cite{he2020pathologycap}. 

Despite this diversity in laguage mdoels fro medical captioning, gastrointestinal endoscopy remains relatively underexplored, with most studies focusing on classification or segmentation, as shown by Aslan et al. \cite{aslan2023colitisai}. The work of Valencia et al. \cite{valencia2023image} ones one of the first works to tackle  the problem of MES grading and captioning in tandem, introducing one of the first captioned datasets for ulcerative colitis exploring different ResNet architectures together with variants of recurrent neural networks, including LSTM and GRU.
More recently,  lesion-aware framework proposed by López-Escamilla et al. (2025) \cite{lopezescamilla2025lesionaware} introduced one of the first captioning pipelines explicitly tailored to ulcerative colitis based on transformers, incorporating localized visual grounding mechanisms to enhance interpretability.

In summary, despite the growing body of work on automated ulcerative colitis analysis, current approaches remain limited in several important aspects. Most existing systems are designed primarily for classification tasks and focus on improving predictive accuracy, while offering limited support for interpretability, semantic reasoning, or clinically meaningful explanations of endoscopic findings. Furthermore, many studies rely on relatively small or single-center datasets, which restricts the ability of models to generalize across variations in imaging equipment, clinical protocols, and patient populations. These limitations highlight the need for more comprehensive evaluation frameworks that combine robust visual analysis with clinically grounded textual interpretation. As discussed in the previous section, advancing toward clinically deployable computer-aided diagnostic systems for ulcerative colitis will require not only improved modeling strategies but also the availability of diverse multi-center datasets and standardized benchmarking platforms. In this context, collaborative challenges and shared datasets as EndoUC can play a pivotal role by enabling systematic comparison of algorithms, encouraging reproducibility, and fostering the development of models that integrate visual understanding with interpretable clinical descriptions.

\section*{Methods}
\subsection*{Dataset}
In this study, a multicenter dataset of gastrointestinal images was gathered from three different sources for the evaluation and assessment of ulcerative colitis. The dataset was designed to improve the variability in imaging characteristics while enhancing the robustness and generalization of deep learning models. Each dataset consists of image level classification and image captions. The image level classification were provided for both Mayo Endoscopic Scoring (MES) from 0 to 3 and UCEIS scoring from 0 to 8. The captions for each of these images were provided based on the characteristics of these images which are detailed below. The curated "Endo-UC" dataset can be divided into three distinct sub-sets based on the source and the image quality.
\paragraph{Sub-set I:} It consisted of 1197 images sourced from previously published works \cite{valencia2023image,xu2022additive}. The dataset consisted of 851 images from publicly available HyperKavsir dataset \cite{Borgli2020,valencia2023image,xu2022additive}, and 131 images of only MES 0 used to tackle class imbalance in \cite{valencia2023image}. Work by Valencia et al. \cite{valencia2023image} for the first time provided a dataset with both MES scoring and captioning, however, it did not consist of UCEIS scoring.
Additionally, 216 were sourced from collected under universal patient consenting at the Translational Gastroenterology
Unit, John Radcliffe Hospital, Oxford) using EVIS Lucera CV260, Olympus Medical Systems \cite{xu2024ssl}. This datasets only included MES scoring. Our team re-annotated these MES scoring and also added UCEIS and captioning which were not present in this dataset. Our Endo-UC dataset for sub-set I is unique and purposeful for multimodal analysis.
%
\paragraph{Sub-set II:} It comprises of 516 images sourced from a clinical center in Egypt, acquired during routine colonoscopy procedures. This subset is captured directly from the endoscopic imaging system using the built-in digital capture of the endoscopy processor, generating a high-resolution (HR) set. The data was collected using Olympus EVIS Medical Systems equipped with a CF-H185L colonoscope. 

\paragraph{Sub-set III:} To further increase the heterogeneity of the dataset and assess the robustness of the model in diverse image settings, this subset consisted of 782 low-resolution (LR) images from the same clinical center in Egypt used for Subset II. These images were obtained by extracting frames from the recorded endoscopic video stream providing additional variability in terms of spatial resolution, lighting settings, and image clarity.  The incorporation of low-resolution dataset enables the assessment of methodologies performance under constrained imaging conditions and reduces potential bias towards high-quality data.

\begin{figure*}
    \centering
    
    \begin{subfigure}[t]{0.48\textwidth}
        \centering
        \includegraphics[width=\linewidth]{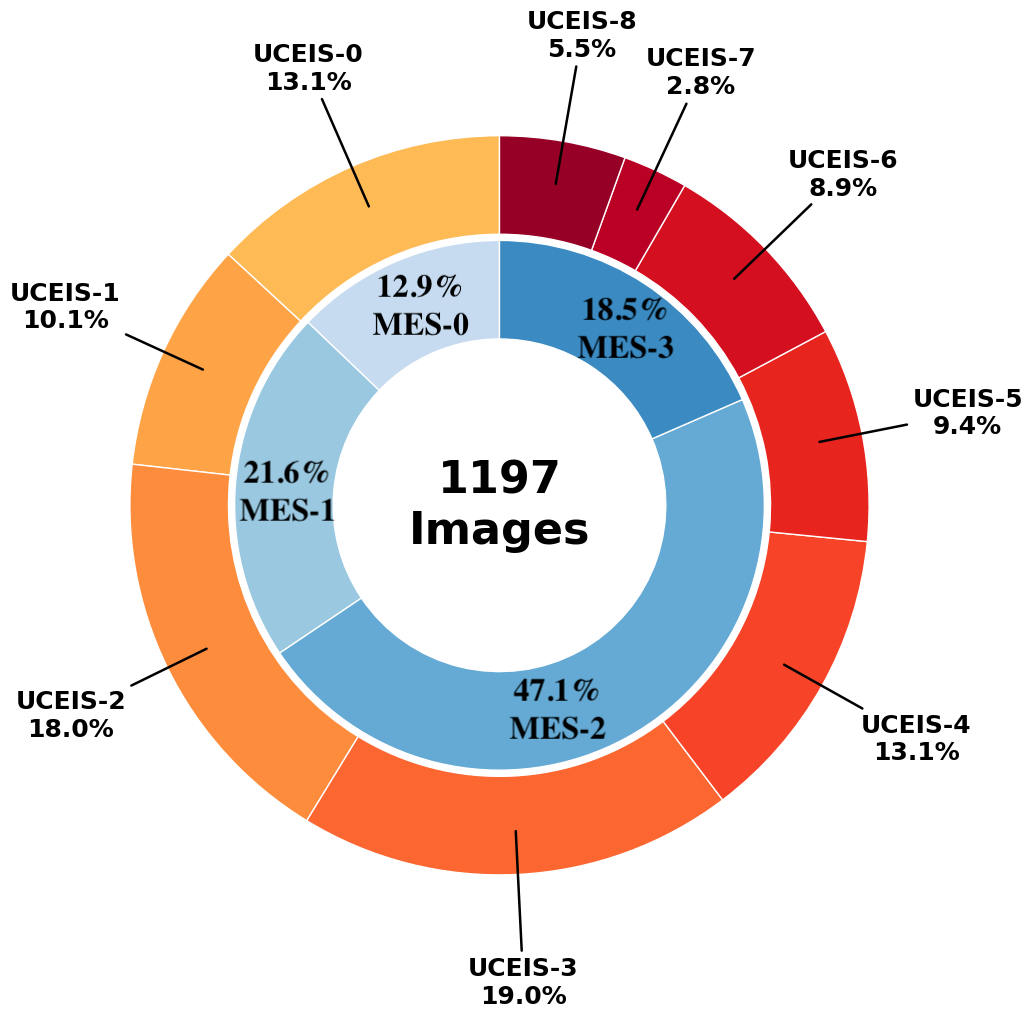}
        \caption{Sub-set I}
    \end{subfigure}
    \hfill
    \begin{subfigure}[t]{0.48\textwidth}
        \centering
        \includegraphics[width=\linewidth]{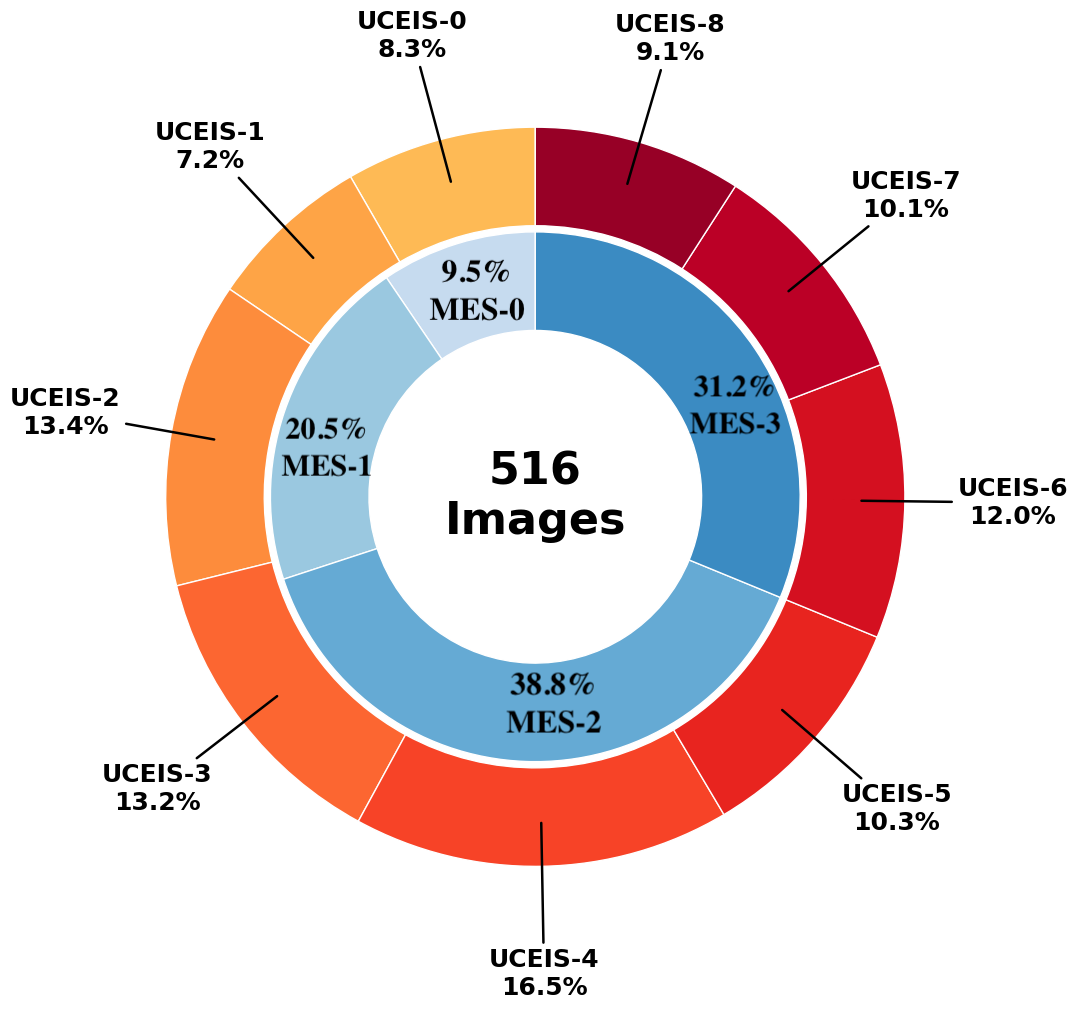}
        \caption{Sub-set II}
    \end{subfigure}
    \hfill
    \begin{subfigure}[t]{0.48\textwidth}
        \centering
        \includegraphics[width=\linewidth]{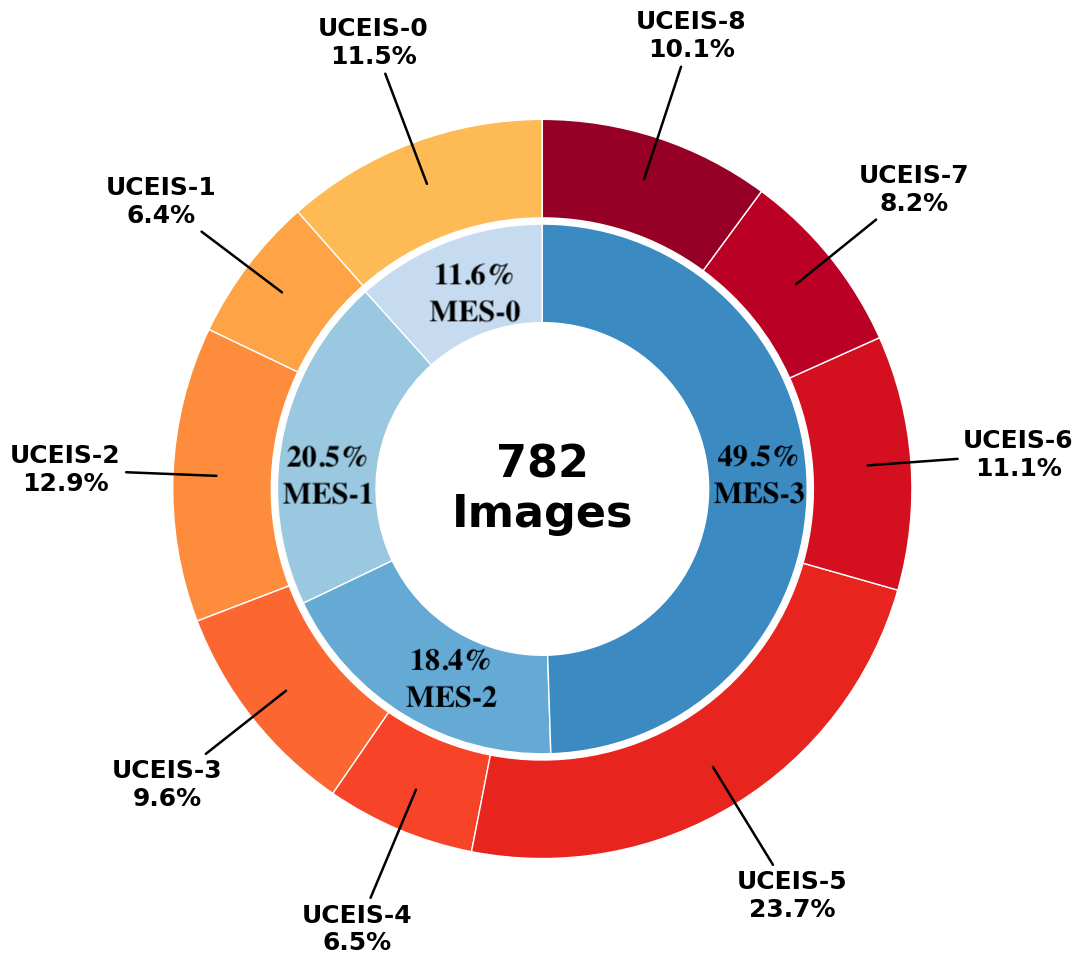}
        \caption{Sub-set III}
    \end{subfigure}
    
    \caption{Distribution of MES and UCEIS scores  categories across three Subsets for the full dataset. The inner ring represents the MES-score distribution while the outer ring represents the UCEIS distribution each with corresponding percentages. The total number of images per subset is shown in the center of the chart}
    \label{fig:three_datasets}
\end{figure*}

\begin{figure}
    \centering
    \includegraphics[width=\textwidth]{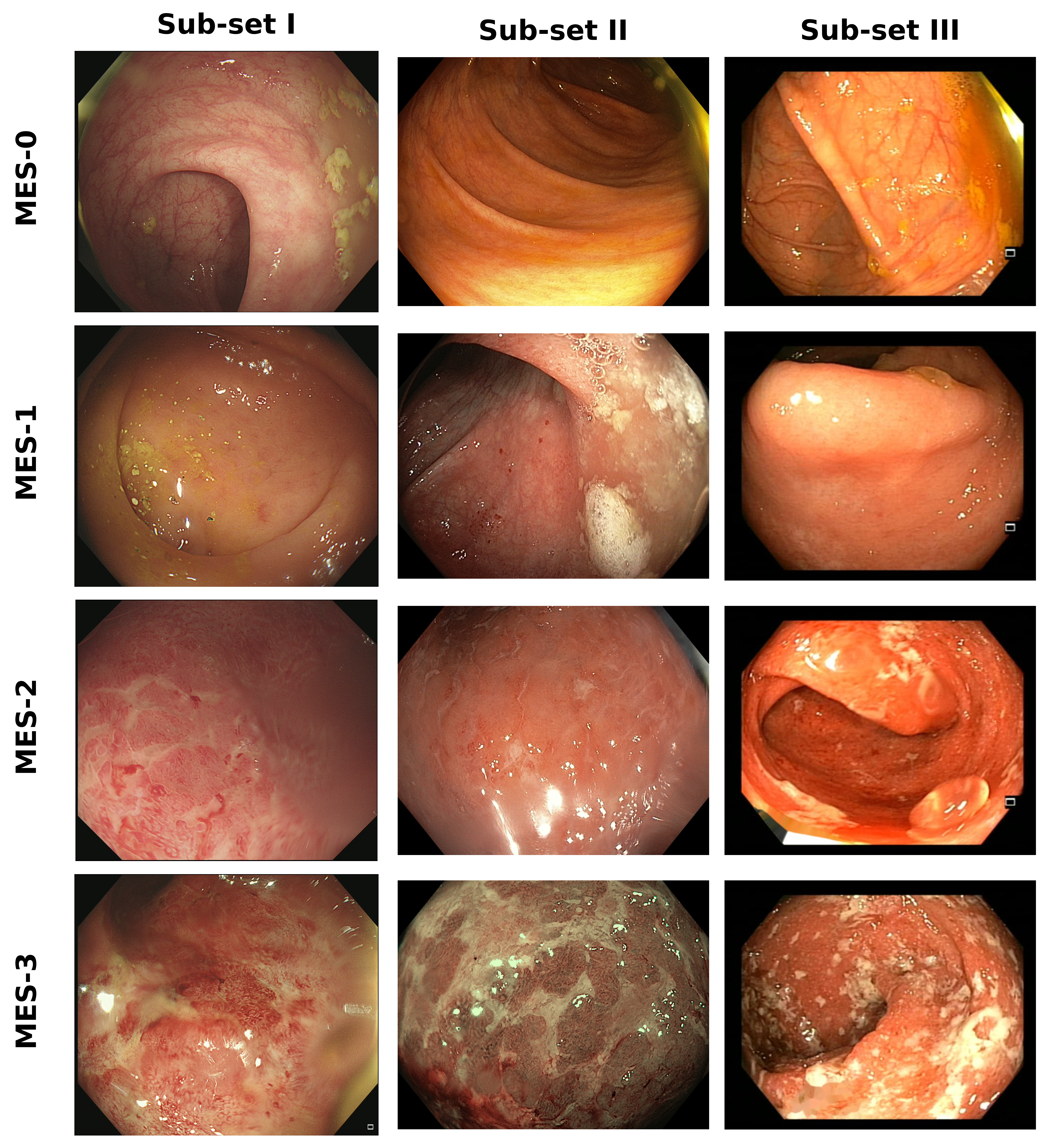}
    \caption{Representative sample for different MES score across the three subsets. Rows presents the MES Scores (MES-0, MES-1, MES-2 and MES-3) while columns presents the three Subsets consecutively}
    \label{fig:FullDataSamples}
\end{figure}

\noindent All images from the three centers were reviewed and annotated by gastroenterologists team to ensure consistency of severity grading classification. Each images was annotated according to Mayo Endoscopic Score (MES), Ulcerative Colitis Endoscopic Index of Severity (UCEIS), vascular pattern, bleeding, erosions and ulcers, friability and erythema. Additionally, we provide a description for each images using the described annotation. The incorporation of the dual scoring systems (MES and UCEIS) allows thorough assessment and supports benchmarking across different clinical frameworks. Figure \ref{fig:three_datasets} presents the distribution of ulcerative colitis severity among the dataset from the different centers, illustrating the correlation between the MES and UCEIS scores. The inner ring presents the percentages of cases within the MES category (i.e. MES-0 to MES-3). In contrast, the outer rings present the corresponding UCEIS distribution (i.e. UCEIS-0 to UCEIS-3) within each MES category, providing the insight for each class. Moreover, the figure highlights the relationship between UCEIS severity and higher MES scores, demonstrating the alignment between the two scoring systems. Furthermore, Table \ref{tab:1} summarizes the distribution of the entire data set of 2495 images illustrating the distribution of the UCEIS grades within each MES category. Additionally, Fig. \ref{fig:FullDataSamples} presents illustrative samples for each MES scores across the three Subsets.

\begin{figure}
    \centering
    \includegraphics[width=\textwidth]{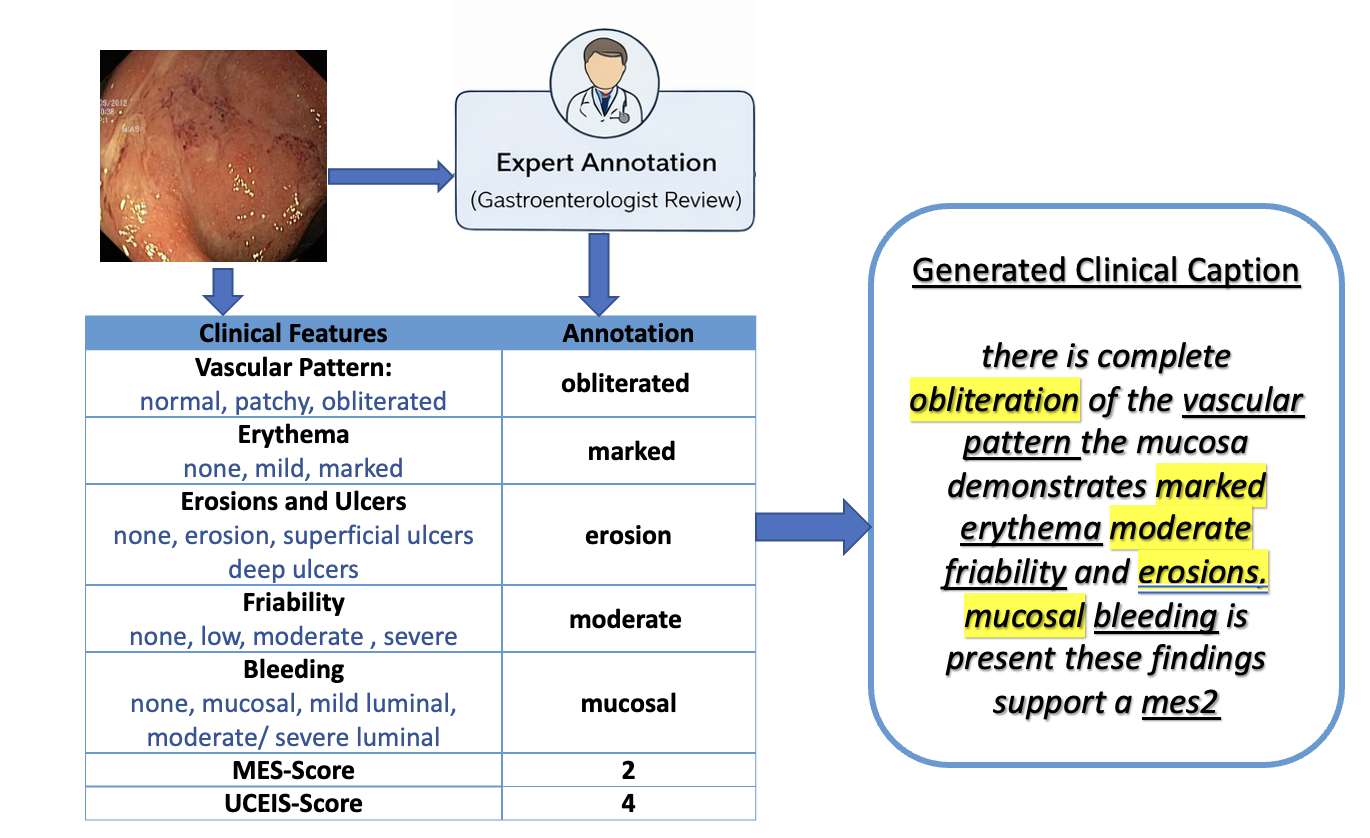}
    \caption{Overview of annotation pipeline, illustrating features selected by gastroenterologist and the generation of descriptive caption.}
    \label{annotation}
\end{figure}

Furthermore, textual captions were generated to provide description for each endoscopic image. During the annotation process, the gastroenterologist annotated each image based on clinically relevant features associated with ulcerative colitis. These features include \textit{vascular pattern} (normal, patchy and obliterated), \textit{bleeding} (none, mucosal, mild luminal and moderate/ severe luminal), \textit{erythema} (none, mild and marked), \textit{erosions and ulcers} (none, erosion, superficial ulcers and deep ulcers) and \textit{friability} (none, low, moderate and severe). These features present the structural and inflammorty changes in the intestinal mucosa and the underlying tissue cause by the ulcerative colitis including mucosal damage, vascular changes, erosion and cell inflammation \cite{taku2020ulcerative}. The annotation of the features allows the gastroenterologist to determine the severity of the ulcerative colitis. In this stage, the severity of these features are evaluated and combined to assign the MES and UCEIS scores for each image \cite{buchner2024aga}. Figure \ref{annotation} presents the overall workflow used to label the full dataset. 

\begin{figure}
    \centering
    \includegraphics[width=\textwidth]{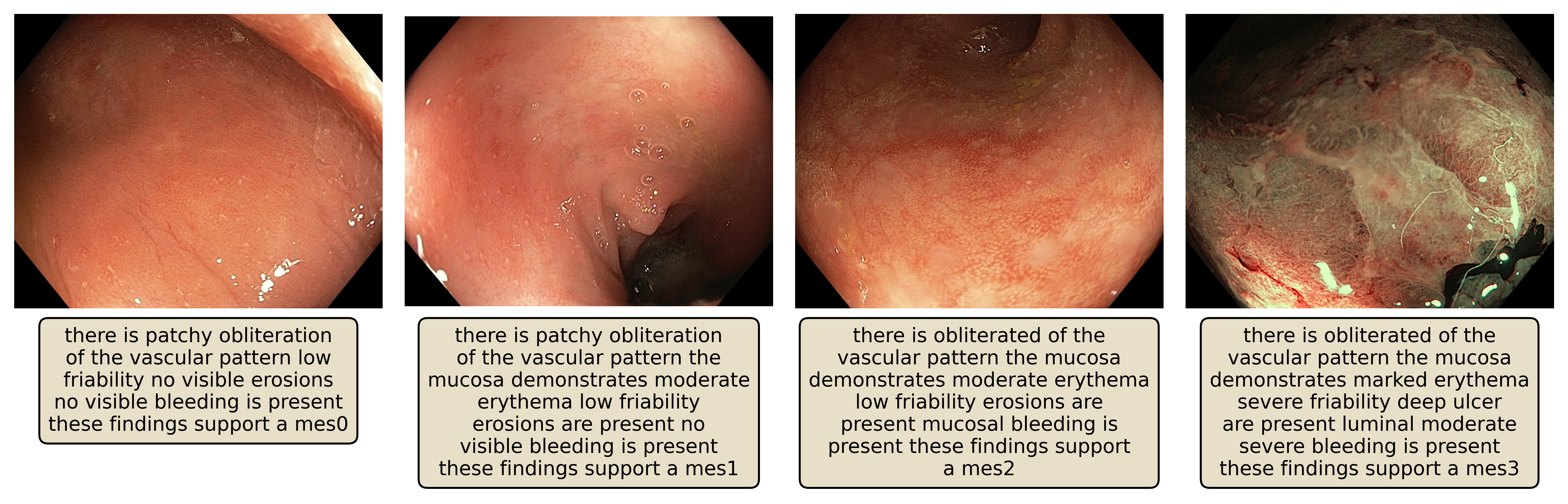}
    \caption{Sample images from the dataset with captions presenting different grades of MES score with varied properties}
    \label{fig:captions}
\end{figure}

Based on these annotations, a textual descriptive caption was produced for each image to summarize the endoscopic findings. To generate these captions, we implemented a python code that combines the properties from these five features into a standardized sentence structure ensuring consistency across the full dataset. Figure \ref{fig:captions} presents sample captions for images for different MES scores with varied UCEIS category.

%

\section*{Technical Validation}
For the technical validation, a total of  1452 frames from two different subsets (Subset I and Subset III) of the proposed data set were used for training baseline models. The evaluation was conducted on 311 images of low-resolution samples for Subset III  which were not included during training phase. The training set was split into 80\% training only and 20\% validation data. 


\begin{table}[t]
\centering
\caption{Performance evaluation of SOTA classification methods of MES scoring on 311 low-resolution test samples similar to the challenge leaderboard. Top two values are presented in bold. Results present, Top-1 Accuracy, F1-Score, AUC: Area Under Curve, Sensitivity, Specificity and  PPV: positive predictive value }
\label{tab:sota_results}

\begin{tabular}{l|cccccc}
\toprule
Model & \textbf{Top-1 Accuracy} & \textbf{F1-score} & \textbf{AUC} & \textbf{Sensitivity} & \textbf{Specificity} & \textbf{PPV} \\
\midrule

\textbf{ResNet50}  \cite{he2016deep} & 0.6013 & 0.4934 & 0.6798 & 0.5101 & 0.8496 & 0.5023 \\
\textbf{SwinTransform}\cite{liu2021swin}  & 0.6720 & 0.5878 & 0.7339 & 0.5953 & 0.8725 & 0.6128 \\
\textbf{DINOv2\_Large} \cite{oquab2023dinov2}& \textbf{0.7138} & \textbf{0.6592} & \textbf{0.7694} & \textbf{0.6477} &\textbf{ 0.8911} & \textbf{0.6746 }\\
\textbf{DINOv2} \cite{oquab2023dinov2} & \textbf{0.6945} & 0.5998 & 0.7448 & 0.6033 & \textbf{0.8863} & \textbf{0.6198} \\
\textbf{DINOv3} \cite{simeoni2025dinov3} & 0.6881 & \textbf{0.6066} &\textbf{ 0.7467} & \textbf{0.6098} & 0.8837 & 0.6126 \\
\textbf{ViT} \cite{dosovitskiy2020image} & 0.6624 & 0.5543 & 0.7289 & 0.5839 & 0.8738 & 0.5707 \\
\textbf{ConvNeXt} \cite{liu2022convnet}   & 0.6624 & 0.5831 & 0.7359 & 0.5976 & 0.8741 & 0.5988 \\

\bottomrule
\end{tabular}
\end{table}

\begin{table*}[t]
\centering
\caption{Performance evaluation of SOTA classification methods for each MES Category (MES-0, MES-1, MES-3 and MES-3) scoring on 311 low-resolution test samples. Top two values are presented in bold. Results present, SEN: Sensitivity, SPE: Specificity and PPV: positive predictive value}
\label{tab:class_results}

\begin{tabular}{l|ccc|ccc|ccc|ccc}
\toprule
\multirow{2}{*}{Model} &
\multicolumn{3}{c}{\textbf{MES-0}} &
\multicolumn{3}{c}{\textbf{MES-1}} &
\multicolumn{3}{c}{\textbf{MES-2} }&
\multicolumn{3}{c}{\textbf{MES-3}} \\
\cmidrule(lr){2-4}\cmidrule(lr){5-7}\cmidrule(lr){8-10}\cmidrule(lr){11-13}
& SEN & SPE & PPV & SEN & SPE & PPV & SEN & SPE & PPV & SEN & SPE & PPV \\
\midrule
\textbf{ResNet50} \cite{he2016deep}
& 0.667 & 0.913 & 0.500
& 0.313 & 0.858 & 0.364
& 0.211 & 0.933 & 0.414
& 0.851 & 0.694 & 0.732 \\

\textbf{SwinTransform}\cite{liu2021swin}
& \textbf{0.806} & 0.942 & 0.644
& 0.375 & \textbf{0.955 }& \textbf{0.686}
& 0.298 & 0.886 & 0.370
& 0.903 & 0.707 & 0.751 \\

\textbf{DINOv2\_Large} \cite{oquab2023dinov2}
& 0.722 & \textbf{0.975} & \textbf{0.788}
& \textbf{0.547} & 0.915 &\textbf{ 0.625}
& \textbf{0.439 }& 0.898 & \textbf{0.490}
& 0.883 & 0.777 & 0.795 \\

\textbf{DINOv2} \cite{oquab2023dinov2}
& 0.667 & \textbf{0.960} &\textbf{ 0.686}
& \textbf{0.609 }& 0.854 & 0.520
& 0.228 & \textbf{0.941} & 0.464
& \textbf{0.909} &\textbf{ 0.790} & \textbf{0.809 }\\

\textbf{DINOv3} \cite{simeoni2025dinov3}
& 0.750 & 0.949 & 0.659
& 0.453 & 0.915 & 0.580
& \textbf{0.333 }& 0.894 & 0.413
& 0.903 & 0.777 & \textbf{0.799} \\

\textbf{ViT} \cite{dosovitskiy2020image}
& \textbf{0.833} & 0.898 & 0.517
& 0.313 & 0.915 & 0.488
& 0.281 & \textbf{0.937} &\textbf{ 0.500}
& \textbf{0.909} & 0.745 & 0.778 \\

\textbf{ConvNeXt} \cite{liu2022convnet}
& 0.778 & 0.927 & 0.583
& 0.328 & \textbf{0.947} & 0.618
& 0.421 & 0.870 & 0.421
& 0.864 & 0.752 & 0.773 \\

\bottomrule
\end{tabular}

\end{table*}

\subsection*{Benchmarking of state-of-the-art methods}
For evaluation purpose we used a set of popular state-of-the-art (SOTA) methods for both classification and captioning tasks. The classification models include \textit{ResNet50} \cite{he2016deep},  \textit{SwinTransform}\cite{liu2021swin}, \textit{DINOv2\_Large} \cite{oquab2023dinov2}, \textit{DINOv2} \cite{oquab2023dinov2}, \textit{DINOv3} \cite{simeoni2025dinov3}, \textit{ViT} \cite{dosovitskiy2020image} and \textit{ConvNeXt} \cite{liu2022convnet}, while captioning models include \textit{BLIP} {\cite{li2022blip}}. All models were trained on GPU was RTX 3500 ADA.


\subsubsection*{Evaluation metrics for classification task.}

We compute standard metrics used for assessing classification performances that includes: \textbf{\textit{Top-1 accuracy}} ($ = \frac{TP+TN} {TP +TN + FP + FN}$), \textbf{\textit{F1-score}} ($ = \frac{2TP} {2TP +FP +FN}$), \textbf{\textit{Sensitivity}} ($SEN={\frac {TP}{TP+FN}}$), \textbf{\textit{Specificity}} ($SPE={\frac {TN}{TN+FP}}$) and \textbf{\textit{Positive predictive value}} ($PPV={\frac {TP}{TP+FP}}$) that are based on  true positives (TP), false positives (FP), true negatives (TN), and false negatives (FN) classification counts. 

\subsubsection*{Evaluation metrics for image captioning task.}
We compute well-known evaluation metrics used to assess captioning taks that include: 
\paragraph{Cosine similarity} measures the similarity between two texts based on the cosine of the angle between their vector representations as follows \textit{Cosine similarity}$=\frac{A \cdot B}{\|A\|\|B\|}$ with $A \cdot B = \sum_{i=1}^{n} a_i b_i$, $\|A\| = \sqrt{\sum_{i=1}^{n} a_i^2}$, $\|B\| = \sqrt{\sum_{i=1}^{n} b_i^2} $.

\paragraph{BLEU (Bilingual Evaluation Understudy)} uses four different $n$-grams and is defined as
$\textit{BLEU} = BP \cdot \exp \left( \sum_{n=1}^{N} w_n \log p_n \right)$,
where $p_n$ denotes the modified $n$-gram precision and $w_n$ are the weights assigned to each $n$-gram. 
The brevity penalty $BP$ is defined as 
$BP = \begin{cases} 1 & \text{if } c > r \\ e^{(1-r/c)} & \text{if } c \le r \end{cases}$,
where $c$ and $r$ denote the candidate and reference caption lengths, respectively.

\paragraph{ROUGE (Recall-Oriented Understudy for Gisting Evaluation)} which is a recall focused metric measured as follows $ROUGE\text{-}N =
\frac{\sum_{gram_n \in Ref} Count_{match}(gram_n)}
{\sum_{gram_n \in Ref} Count(gram_n)}$ where $gram_n=$ n-grams in the reference caption and $count_{match}=$ number of overlapping n-grams.

\paragraph{METEOR (Metric for Evaluation of Translation with Explicit ORdering)} measures the average unigram recall and precision where $F_{mean} = \frac{10PR}{R + 9P}$ and $METEOR = F_{mean}(1 - Penalty)$ with $Penalty = 0.5 \left(\frac{chunks}{matches}\right)^3$  where chuncks are the number of contiguous matched segments and matches are the number of matched unigrams. 

\begin{table}[t]
\centering
\caption{Performance evaluation of SOTA captioning methods on 311 low-resolution test samples. }
\label{tab:caption_results}

\begin{tabular}{l|cccccc}
\hline
\textbf{Model} & \textbf{Cosine Similarity} & \textbf{BLEU-1} & \textbf{BLEU-4} & \textbf{METEOR} & \textbf{ROUGE-1} & \textbf{ROUGE-L} \\
\hline
\textbf{BLIP-Base \cite{li2022blip}} & 0.7590 & 0.7887 & 0.6573 & 0.8124 & 0.8338 & 0.8298 \\
\textbf{BLIP-Large \cite{li2022blip} }& \textbf{0.7977} & \textbf{0.8210} & \textbf{0.6960} & \textbf{0.8435} & \textbf{0.8615} & \textbf{0.8574} \\
\hline
\end{tabular}

\end{table}

\subsection*{Validation Summary}
Table \ref{tab:sota_results} shows the overall performance of the various SOTA models for MES scoring on the provided test data composed of 311 images from subset II. Our experimental evolution suggests that the DINOv2\_Large achieves the best overall performance across several evaluation metrics for MES classification, obtaining a Top-1 Accuracy (0.7138), F1-Score (0.6592), AUC (0.7694), Sensitivity (0.6477), Specificity (0.8911) and PPV (0.6746). These findings suggest that the self-supervised visual representation of features learned by the DINOv2 architecture provides improved feature representation when compared to the other models.

The DINOv2 and DINOv3 models also demonstrated a competitive performance, where the DINOv2 achieved the second-best performing Top-1 Accuracy (0.6945) while the DIN0v3 obtained the second-highest F1-Score (0.6066). Additionally, both models had better performing results in terms of AUC, sensitivity, specificity, and PPV when compared to the rest of the baseline models. Transformer-based models such as SwinTransform and ViT showed moderate performance across most metrics, with a Top1-Accuracy (0.6720) and (0.6224), respectively. On the other hand, conventional architectures such as ResNet-50 and ConvNext achieved lower performance, suggesting that these models may have a limited ability to capture the complex mucosal variation present in ulcerative colitis.  

While Table \ref{tab:sota_results} provides an overall classification performance, a more detailed analysis across the four severity levels is shown in Table \ref{tab:class_results}. For \textbf{MES-0}, the ViT achieved the highest sensitivity (0.833) while the DINOv2\_Large achieved the highest specificity (0.975) and PPV (0.788), demonstrating effectiveness in classifying non-inflamed mucosa, which is very challenging. For \textbf{MES-1}, which presents mild inflammatory patterns, DINOv2 demonstrated the highest sensitivity (0.609). However, the SwinTransform improved its ability in classifying the category with top-performing specificity and PPV values (0.955) and (0.686), respectively. The performance among models shows moderate variation, which reflects feature differences associated with the intermediate severity level. For \textbf{MES-2}, the results demonstrate similar performance between models, where DINOv2\_Large achieved the highest sensitivity (0.439), while DINOv2 provided the highest specificity (0.941). Moreover, the ViT obtained the highest PPV (0.500). Finally, the DINOv2 had an outstanding performance when classifying the  \textbf{MES-3}, which represents the most severe stage of ulcerative colitis with top scores for the sensitivity (0.909), specificity (0.790) and PPV values (0.809).  

Furthermore, Table \ref{tab:caption_results} provides the quantitative evaluation of the captioning SOTA models on the test data. The BLIP-Large model maintained high performance across all the evaluation measures, with cosine similarity (0.7977) showing a robust similarity alignment between the generated caption and ground-truth. In addition to the rest of the evaluation metrics, BLEU-1 (0.8210), BLUE-4 (0.6960), METEOR (0.8435), ROUGE-1(0.8615), and ROUGE-L (0.8574).

%
\section*{Discussion}
In this paper, we present a dataset and evaluation of baseline methods for the analysis of ulcerative colitis from endoscopic images. This includes the classification of ulcerative colitis severity based on the MES-score and captioning generation to describe relevant mucosal features. For the classification evaluation, the DINOv2 models showed superior performance when compared to different SOTA models. These overall results of sensitivity, specificity and PPV were obtained by averaging the performance across all MES-scores. In terms of overall evaluation, the DINOV2\_Large outperformed the other models for all evaluation metrics. These findings suggest that the self-supervised based models can detect challenging visual patterns related to mucosal inflammation more effectively. This is an important feature for classifying the ulcerative colitis as the differentiation between severity levels mainly depends on fine texture details and vascular patterns. 

The per-class evaluation demonstrates that the model's performance varies among the different MES categories. As shown, the MES-3 (i.e., severe ulcerative colitis category) generally achieves high sensitivity across most of the models, which indicates that advanced stages can provide more distinctive properties that are easier for models to detect. However, MES-2 and MES-1 (i.e., moderate ulcerative colitis severity) appear to be more challenging to classify accurately, with sensitivities of 0.439 and 0.609, respectively. The difference between these two mild severities can be challenging to classify due to the subtle difference in mucosal texture and vascular patterns.

In addition to MES classification, we investigate the generations of textual captions to describe image severity in terms of clinical features. As shown in Table \ref{tab:caption_results} we compared the two versions of BLIP model where the BLIP-Large outperformed BLIP-Base in all evaluation metrics. The enhanced performance BLIP-Large is due to its larger model capacity and better multi-modal representation learning enabling improved alignment between visual features and text descriptors . The ability of generating captions describing endoscopic images in an essential tool that can assist clinicians by providing mucosal findings automatically in order to support reporting and decision-making process. Moreover, integrating merging image classification with textual captioning generation may provide useful frameworks where models can provide both severity prediction and describe visual features that contributes to predictions.

\bibliography{sample}

%
%

\end{document}